%% file: iclr2027_conference.tex
\documentclass{article} % For LaTeX2e
\usepackage{iclr2027_conference,times}

\input{math_commands.tex}

\usepackage{xcolor}
\definecolor{navyblue}{RGB}{0,0,128}
\definecolor{Darkklein}{HTML}{122A82}
\usepackage{hyperref}
\hypersetup{
    colorlinks=true,
    citecolor=navyblue,
    linkcolor=black,
    urlcolor=black
}
\usepackage{url}
\usepackage[capitalize]{cleveref}
\usepackage{booktabs}
\usepackage{amsthm}
\usepackage{graphicx}

\usepackage{float}
\usepackage{comment}

\title{Should We Skip Diffusion?}

\author{
Yiping Ji$^{1}$\thanks{Corresponding author: \texttt{yipingji96@gmail.com}.}
\quad James Martens$^{2}$ \quad Simon Lucey$^{1}$ \\
$^{1}$Australian Institute for Machine Learning, Adelaide University \\
$^{2}$Independent Researcher
}

\iclrfinalcopy % Uncomment for camera-ready version, but NOT for submission.
\begin{document}

\maketitle

\input{sec/0_abstract}

\input{sec/1_intro}

\input{sec/2_literature}

\input{sec/3_preliminary}

\input{sec/4_method}

\input{sec/5_experiment}

\input{sec/6_conclusion}

\newpage

\bibliography{iclr2027_conference}
\bibliographystyle{iclr2027_conference}

\newpage
\input{sec/7_appendix}

\end{document}

%% file: math_commands.tex
\usepackage{amsmath,amsfonts,bm}

\def\eqref#1{equation~\ref{#1}}
\def\1{\bm{1}}

\DeclareMathAlphabet{\mathsfit}{\encodingdefault}{\sfdefault}{m}{sl}
\SetMathAlphabet{\mathsfit}{bold}{\encodingdefault}{\sfdefault}{bx}{n}

%% file: sec/0_abstract.tex
\begin{abstract}

\begin{comment}
Diffusion transformers learn abstract representations while generating images, but abstraction and the preservation of detail for generation place different demands on the network. The Decoupled Diffusion Transformer (DDT) separates these roles into a condition encoder and a velocity decoder. However, skip/residual connections in the encoder allow shallow features to bypass successive transformations, which may limit progressive abstraction. We propose DDT-RFE, which removes the residual connections around the Self-Attention and MLP operations in each encoder block while maintaining stable training. To retain the information that abstraction discards but that the decoder still needs, we fuse the input patch embedding with intermediate and final encoder features to form the encoder output. The decoder thus has access to information from multiple encoder depths, while each encoder block is able to learn more abstract representations. DDT-RFE achieves overall improvements over DDT across visual understanding tasks, including image classification, semantic segmentation, object discovery, and semantic correspondence, while using fewer encoder blocks. It also achieves a lower FID for image generation on ImageNet.
\end{comment}

Diffusion models learn semantic representations while generating images. In the Decoupled Diffusion Transformer (DDT), a condition encoder provides features that guide a velocity decoder in denoising. To enable effective denoising at all noise levels, these features must capture both high-level abstract structures and low-level details. However, skip/residual connections in the encoder allow shallow features to bypass successive transformations, which may limit progressive abstraction, or at least make it difficult to disentangle different levels of abstraction. We propose DDT-RFE, which removes the residual connections around the Self-Attention and MLP operations in each encoder block while maintaining stable training. To retain the information that abstraction discards but that the decoder still needs, we fuse the input patch embedding with intermediate and final encoder features to form the encoder output. The decoder thus has access to information from multiple encoder depths, while each encoder block is able to learn more abstract representations. DDT-RFE achieves overall improvements over DDT across visual understanding tasks, including image classification, semantic segmentation, object discovery, and semantic correspondence, while using fewer encoder blocks. It also achieves a lower FID for image generation on ImageNet.
\end{abstract}

%% file: sec/1_intro.tex
\section{Introduction}

Diffusion transformers~\citep{dit} have emerged as a dominant backbone for image generation, achieving remarkable fidelity and diversity across a wide range of visual domains. Beyond synthesis, their intermediate activations encode semantic information that supports downstream visual recognition and dense prediction tasks~\citep{stracke2025cleandift,gan2025unleashing, helbling2025conceptattention,xiang2023denoising}. Crucially, representation learning and image synthesis are not independent processes: in standard diffusion transformers, semantic abstraction and fine-grained signal reconstruction are implicitly coupled within the same denoising network. Recent work has begun to explicitly decouple these two roles~\citep{yu2026pixeldit,Zheng2024MaskDiT}. Decoupled Diffusion Transformer (DDT)~\citep{wang2026ddt}, for example, introduces a ``condition encoder" for learning representations and a ``velocity decoder" for signal reconstruction. The condition encoder retains the standard diffusion transformer block design and transforms noisy inputs into representations that provide the decoder with information useful for reconstruction, including abstract structures and low-level image details.

\begin{figure}
    \centering
    \includegraphics[width=\linewidth]{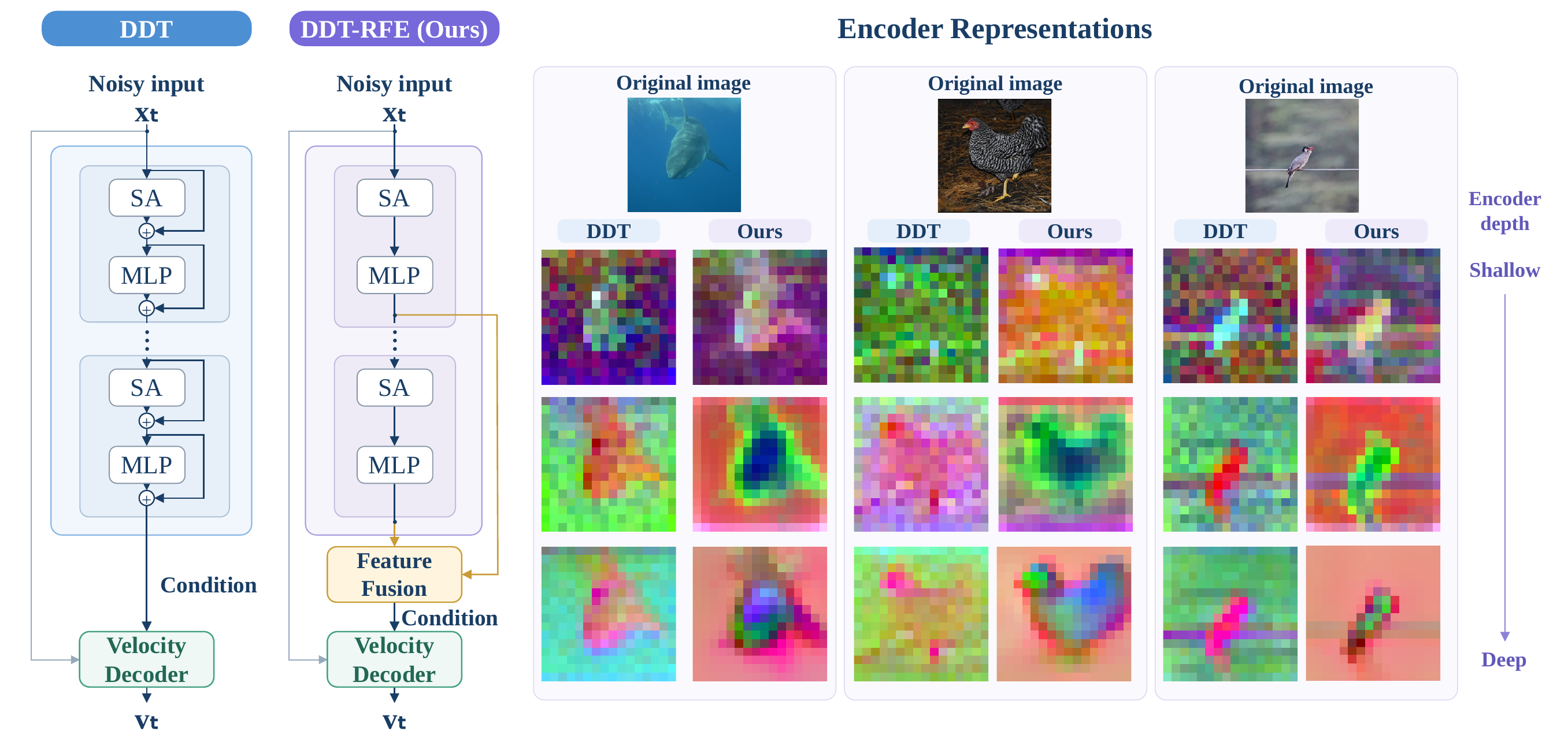}
    \caption{
        \textbf{Left:} Overall architectures of DDT and DDT-RFE (our method).
        \textbf{Right:} Encoder representations from shallow to deep layers at timestep $t=0.5$.
        Features are projected into three dimensions using PCA and visualized as RGB images.
        Our method produces more abstract representations with clean backgrounds and salient objects. More visualizations are shown in ~\cref{fig:ddt_pca_t0.1,fig:ddt_pca_t0.3,fig:ddt_pca_t0.5}
    }
    \label{fig:teaser}
\end{figure}

However, how the encoder develops abstract features and combines features at different levels affect the quality of the representations that are provided to the decoder. 
\begin{comment}
In DDT, residual connections (or skip connections)~\citep{he2016deep} couple these two roles. They implicitly aggregate information across depth, allowing different levels of features to persist alongside successive transformation~\citep{raghu2021vision}.    
\end{comment}
In DDT, each encoder block transforms its input and integrates the resulting features with the input through residual connections~\citep{he2016deep}. Stacking blocks allows features to undergo successive transformations, which supports abstract representations. At the same time, residual connections carry information from earlier blocks into the inputs of later transformations. Although this can retain information useful for reconstruction, we hypothesize that the repeated accumulation of shallow features may hinder  abstraction from the noisy input~\citep{ji2026cutting,Zhang_2026_CVPR}. As shown in \cref{fig:teaser}, object structure remains difficult to distinguish in the visualized DDT encoder features, even in deeper blocks. This raises a central question: can the encoder develop more abstract features through successive transformations, while combining them with earlier features at its output, to provide the decoder with information at multiple levels of abstraction?

To address this, we propose DDT with a Residual-Free Encoder (DDT-RFE), which combines a residual-free condition encoder with explicit fusion of intermediate and final encoder features to condition the decoder. We remove the residual connections around the Self-Attention and MLP operations in each encoder block, eliminating the identity shortcuts through which earlier features bypass these transformations. This design encourages progressive abstraction through successive transformations. In order to improve the convergence of the residual-free encoder, we propose an initialization scheme inspired by \citet{ji2026cutting} but with fewer hyperparameters to tune. To retain access to information that is lost during abstraction but may still be needed for reconstruction, we fuse the input patch embedding and intermediate encoder features with the final encoder block output to form the decoder's condition. The decoder can therefore access representations at different levels, without requiring all useful information to persist in the final encoder block representation.

We evaluate DDT-RFE on four visual tasks and image generation. The results show that when DDT-RFE's encoder is ultilized for visual tasks, it achieves overall improvements in representation quality while remaining efficient with nearly 40\% fewer encoder blocks. On ImageNet at $256\times256$ resolution, DDT-RFE reduces FID from 8.98 to 7.68 at the L scale and from 7.27 to 6.81 at the XL scale under the same training budget.
Our main contributions are as follows:
\begin{itemize}
    \item We propose DDT-RFE, which combines a residual-free condition encoder, a new initialization scheme, and feature fusion to improve representation learning and image generation.
    \item Qualitatively, our feature visualizations show clearer object structure and cleaner backgrounds in the representations learned by DDT-RFE than in those learned by DDT.
    \item Quantitatively, DDT-RFE achieves overall improvements over DDT on image classification, semantic segmentation, object discovery, and semantic correspondence with fewer encoder blocks. It also achieves a lower FID for image generation on ImageNet.
\end{itemize}

%% file: sec/2_literature.tex
\section{Related Work}
\paragraph{Diffusion transformer architectures.}
DiT~\citep{dit} establishes a scalable backbone for latent diffusion
built from a stack of transformer blocks. U-DiT~\citep{tian2024u} adds
long skip connections that carry shallow features to deeper layers.
Other designs separate the learning of semantic representations from
reconstruction. In pixel-space diffusion, PixelDiT~\citep{yu2026pixeldit}
models global semantics at the patch level in an encoder and refines
details at the pixel level in a decoder. DDT~\citep{wang2026ddt} uses a
condition encoder to learn representations that condition a velocity
decoder for signal reconstruction. We build on the condition encoder of
DDT so that the representations learned in the encoder can be studied in isolation from the decoder.
\paragraph{Representation learning with diffusion models.}
Diffusion denoising has been studied as a self-supervised objective
for learning visual representations that transfer effectively to
downstream tasks.  Through pretraining on image
generation, \citet{xiang2023denoising} demonstrate that diffusion
models learn linearly separable intermediate representations and achieve image classification performance comparable to that of contemporary contrastive methods.
For diffusion transformers,
DiTF~\citep{gan2025unleashing} investigates the capabilities of DiTs
for accurate dense correspondence and proposes a training-free
method to extract semantically discriminative features from DiTs,
achieving superior performance across multiple benchmarks.
Furthermore, CleanDIFT~\citep{stracke2025cleandift} introduces a
lightweight, unsupervised fine-tuning method that enables diffusion
backbones to provide high-quality, noise-free semantic features
for downstream visual tasks.
Our work examines how the condition encoder shapes representations,
with a focus on progressive abstraction, and how information that
may be discarded during this process can remain accessible to the
decoder for reconstruction.
\begin{comment}
Improved representations can, in turn, benefit generation:
REPA~\citep{yu2025representation} aligns diffusion representations
with pretrained visual features to improve training efficiency
and generation quality.
\end{comment}
\begin{comment}
\paragraph{Residual connections in transformers.}
Residual connections affect both optimization and the representations
that transformers learn.
\citet{ji2025always} link the training difficulties caused by removing
attention shortcuts to poor conditioning of self-attention operation,
and analyze how residual pathways improve their network conditioning.
To address this optimization challenge, subsequent work develops
initialization and optimization strategies that enable effective
training of residual-free
transformers~\citep{ji2026cutting,ji2026quantization}.
Beyond optimization, residual pathways allow information from earlier
layers to persist in deeper representations~\citep{raghu2021vision}. In generative models, \citet{Zhang_2026_CVPR} attenuate residual
connections, including in diffusion transformers, and use long
encoder--decoder connections to retain information for reconstruction.
\citet{ji2026cutting} further report that residual-free vision
transformers learn more abstract representations which benefits dense prediction tasks.
However, while discarding some low-level details may benefit dense prediction, these details may still be needed for reconstruction in diffusion models. 
\end{comment}
\paragraph{Residual connections in transformers.}
Residual connections affect both optimization and the learned representations.
\citet{ji2025always} link the training difficulties caused by removing
attention shortcuts to poor network conditioning in self-attention and
analyze how residual pathways improve network conditioning.
To address this optimization challenge, subsequent work develops
initialization and optimization strategies that enable effective
training of residual-free
transformers~\citep{ji2026cutting,ji2026quantization}.
Beyond optimization, residual pathways allow information from earlier
blocks to persist in deeper representations~\citep{raghu2021vision,Zhang_2026_CVPR}.
\citet{ji2026cutting} further report that residual-free vision
transformers learn more abstract representations and achieve improved
performance on dense prediction tasks. An additional consideration in our setting is how to include more
abstract features in the representations provided to the decoder
while retaining other information useful for reconstruction.
Our work examines this question in DDT by removing residual
connections within the condition encoder and aggregating features
from different depths at its encoder output.

%% file: sec/3_preliminary.tex
\section{Preliminaries}

\subsection{Decoupled Diffusion Transformer}
\label{sec:ddt-preliminaries}

DDT~\citep{wang2026ddt} provides a dedicated condition encoder for semantic extraction and a velocity decoder for signal reconstruction, rather than relying on homogeneous stacks of DiT/SiT blocks. This explicit separation makes DDT particularly suitable for analyzing how information flow within the condition encoder affects representation learning. 

DDT consists of a condition encoder $E_{\phi}$ and a velocity
decoder $D_{\theta}$, both operating in the latent space of a
pretrained Variational Autoencoder
(VAE)\footnote{\url{https://huggingface.co/stabilityai/sd-vae-ft-ema}}.
The VAE encoder maps each image to a latent representation,
which we write in vectorized form as
$\mathbf{x}_0 \in \mathbb{R}^{d_x}$.

The noisy latent $\mathbf{x}_t$ and noise $\boldsymbol{\epsilon}$ have the same dimension and a noisy latent is constructed as
\begin{equation}
    \mathbf{x}_t = \alpha_t\mathbf{x}_0
    + \sigma_t\boldsymbol{\epsilon},
    \qquad
    \boldsymbol{\epsilon}\sim\mathcal{N}(\mathbf{0},\mathbf{I}),
    \label{eq:noisy-latent}
\end{equation}
where $t\in[0,1]$ is the diffusion timestep, $\alpha_t = 1-t$ and $\sigma_t = t$
specify the signal and noise coefficients so that $\mathbf{x}_0$ is
the clean latent and $\mathbf{x}_1$ is pure noise.
The encoder maps the noisy latent $\mathbf{x}_t$ to a
representation under timestep and class conditioning.
This representation is combined with the timestep embedding
to condition the decoder, which takes $\mathbf{x}_t$
as its input:
\begin{align}
    \mathbf{Z}_t &= E_{\phi}(\mathbf{x}_t; t,y),
    \label{eq:ddt-encoder} \\
    \hat{\mathbf{v}}_t &= D_{\theta}(\mathbf{x}_t; \mathbf{Z}_t,t).
    \label{eq:ddt-decoder}
\end{align}
where $y$ is the class label and
$\mathbf{Z}_t\in\mathbb{R}^{N\times D}$ is the encoder output
feature matrix, with $N$ tokens and feature dimension $D$. We focus on the condition encoder to examine how residual connections shape the representations it learns.

% In the original DDT architecture, the class information reaches
% the decoder through $\mathbf{Z}_t$.
% The encoder receives supervision from the decoder's velocity
% prediction loss, together with a representation alignment objective (REPA)~\citep{yu2025representation}. To isolate the effect of the encoder architecture on representation learning, we discard the REPA objective in our experiments and supervise the encoder through the decoder's velocity prediction loss only.

\subsection{Condition Encoder Architecture}
\label{sec:encoder-architecture}
The noisy latent is partitioned into non-overlapping patches and each patch is flattened and linearly projected to a $D$-dimensional token, yielding $\mathbf{H}_0\in\mathbb{R}^{N\times D}$, where $N$ is the number of patches. 
Let $\mathbf{H}_\ell\in\mathbb{R}^{N\times D}$ denote the output
of block $\ell$, for $\ell=1,\ldots,L$. For clarity, we omit timestep $t$.

\paragraph{Conditioning.}
Timestep and class information are injected through adaptive layer normalization
(AdaLN)~\citep{dit}. A conditioning vector
$\mathbf{c} = \mathrm{SiLU}\big(e_t(t) + e_y(y)\big) \in \mathbb{R}^D$
is mapped by a per-block affine layer to shift, scale, and gate
vectors $\boldsymbol{\beta}_i^\ell, \boldsymbol{\gamma}_i^\ell,
\mathbf{g}_i^\ell \in \mathbb{R}^D$ for the attention ($i=1$) and MLP
($i=2$). Here, $e_t(\cdot)$ and $e_y(\cdot)$ denote the timestep embedding function and the class embedding function, respectively. 

\paragraph{Residual Encoder Block.}
Each encoder block applies the following residual updates:
\begin{align}
\tilde{\mathbf{H}}_\ell
&=
\mathbf{H}_{\ell-1}
+
\mathbf{g}_1^\ell \odot
\mathcal{A}_\ell(\mathbf{H}_{\ell-1}),
\label{eq:encoder-sa}
\\
\mathbf{H}_\ell
&=
\tilde{\mathbf{H}}_\ell
+
\mathbf{g}_2^\ell \odot
\mathcal{M}_\ell(\tilde{\mathbf{H}}_\ell),
\label{eq:encoder-mlp}
\end{align}
where the Self-Attention and MLP transformations are defined as
\begin{equation}
\begin{aligned}
\mathcal{A}_\ell(\mathbf{H})
&=
\mathrm{SA}_\ell\!\left(
\mathrm{Norm}_{1}^{\ell}(\mathbf{H})
\odot(\mathbf{1}+\boldsymbol{\gamma}_1^\ell)
+\boldsymbol{\beta}_1^\ell
\right),
\\
\mathcal{M}_\ell(\mathbf{H})
&=
\mathrm{MLP}_\ell\!\left(
\mathrm{Norm}_{2}^{\ell}(\mathbf{H})
\odot(\mathbf{1}+\boldsymbol{\gamma}_2^\ell)
+\boldsymbol{\beta}_2^\ell
\right).
\end{aligned}
\label{eq:sublayers}
\end{equation}
Here, $\odot$ denotes element-wise multiplication,
with channel-wise modulation vectors broadcast across tokens.

The $\mathrm{SA}_\ell$ operation uses four learnable projection matrices,
$\mathbf{W}_{\mathrm{Q}}^\ell,\mathbf{W}_{\mathrm{K}}^\ell,
\mathbf{W}_{\mathrm{V}}^\ell,\mathbf{W}_{\mathrm{O}}^\ell
\in\mathbb{R}^{D\times D}$, corresponding to the query, key, value,
and output projections, respectively.
The $\mathrm{MLP}_\ell$ operation uses two learnable weight matrices,
$\mathbf{W}_1^\ell\in\mathbb{R}^{D_{\mathrm{ff}}\times D}$ and
$\mathbf{W}_2^\ell\in\mathbb{R}^{D\times D_{\mathrm{ff}}}$,
where $D_{\mathrm{ff}}$ is the MLP hidden dimension.
Detailed definitions of the $\mathrm{SA}$ and $\mathrm{MLP}$ operations
are provided in \cref{app:sa-mlp}.

Simply, we combine the residual updates within block $\ell$ into a function
$f_\ell$. The block output can then be written as
\begin{equation}
    \mathbf{H}_\ell = \mathbf{H}_{\ell-1}
    + f_\ell(\mathbf{H}_{\ell-1}),
    \label{eq:residual-block}
\end{equation}

\paragraph{Encoder Output.}
The final encoder output, $\mathbf{Z}_t = \mathbf{H}_L$, is supplied
to the decoder. Expanding the residual updates gives
\begin{equation}
\begin{aligned}
\mathbf{Z}_t
&= \mathbf{H}_0 + f_1(\mathbf{H}_0)
+ f_2\!\left(\mathbf{H}_0 + f_1(\mathbf{H}_0)\right) + \cdots
+ f_L\!\left(
    \mathbf{H}_0
    + \sum_{j=1}^{L-1} f_j(\mathbf{H}_{j-1})
\right).
\end{aligned}
\label{eq:residual-sum}
\end{equation}
This expansion shows that shallow features persist as the encoder blocks go deeper.

%% file: sec/4_method.tex
\section{Method}
\label{sec:residual-free-encoder}
In this section, we introduce our approach to improving the representations learned by the DDT encoder, including a residual-free encoder architecture and initialization strategies. 

\paragraph{Residual-Free Encoder}

We remove the residual connections around the attention and
MLP sublayers within each encoder block while retaining
the timestep- and class-dependent modulation.
Using $\mathcal{A}_\ell$ and $\mathcal{M}_\ell$ defined in
\cref{eq:encoder-sa,eq:encoder-mlp}, each block computes
\begin{align}
\tilde{\mathbf{H}}_\ell
&=
(\mathbf{1}+\boldsymbol{\mathbf{g}}_1^\ell)\odot
\mathcal{A}_\ell(\mathbf{H}_{\ell-1}),
\label{eq:residual-free-sa}
\\
\mathbf{H}_\ell
&=
(\mathbf{1}+\boldsymbol{\mathbf{g}}_2^\ell)\odot
\mathcal{M}_\ell(\tilde{\mathbf{H}}_\ell).
\label{eq:residual-free-mlp}
\end{align}
We parameterize the output gates as
$\mathbf{1}+\boldsymbol{\mathbf{g}}_1^\ell$ and
$\mathbf{1}+\boldsymbol{\mathbf{g}}_2^\ell$.
We initialize the gate parameters to zero so that both output gates
equal $\mathbf{1}$ at initialization.

Let $f_\ell^{\mathrm{RF}}$ denote the gated transformation of residual-free encoder block $\ell$. In contrast to \cref{eq:residual-block}, each residual-free encoder block is computed as $\mathbf{H}_\ell = f_\ell^{\mathrm{RF}}(\mathbf{H}_{\ell-1})$, without additive identity pathways. We expect these successive transformations to encourage progressive semantic abstraction.

\paragraph{Encoder Output.}
To provide the decoder with both detailed and abstract information from multiple encoder depths,
we aggregate the input patch embeddings and the outputs of all
encoder blocks:
\begin{equation}
\mathbf{Z}_t = \sum_{\ell=0}^{L} \mathbf{H}_{\ell},
\label{eq:fused-condition}
\end{equation}
where $\mathbf{H}_0$ denotes the input patch embeddings and
$\mathbf{H}_{\ell}$ denotes the output of encoder block $\ell$.
In contrast to the residual accumulation in
\cref{eq:residual-sum}, this aggregation is performed only
at the encoder output, and does not introduce identity pathways
within the encoder.
It gives the decoder direct access to features from different
depths, without requiring deeper encoder blocks to retain all
the information needed for reconstruction.

% \paragraph{Encoder Output.}

% To provide the decoder with both detailed and abstract information from multiple encoder depths, we aggregate the input patch embeddings and encoder block outputs:
% % \begin{equation}
% % \mathbf{Z}_t = \mathbf{H}_0 + \sum_{\ell\in\mathcal{S}}\mathbf{H}_\ell,
% % \label{eq:fused-condition}
% % \end{equation}
% \begin{equation}
% \mathbf Z_t
% =
% \mathbf H_0+
% \sum_{\ell\in\mathcal S}
% f_\ell^{\mathrm{RF}}\!\left(f_{\ell-1}^{\mathrm{RF}}\!\left(\cdots f_1^{\mathrm{RF}}(\mathbf H_0)\cdots\right)\right),
% \label{eq:fused-condition}
% \end{equation}
% where $\mathcal{S}\subseteq\{1,\ldots,L\}$ denotes the selected encoder depths and includes the final block. We use the outputs of all encoder blocks, i.e., $\mathcal{S}=\{1,\ldots,L\}$. With this choice, the encoder output can be expanded as follows:
% \begin{equation}
% \begin{aligned}
% \mathbf{Z}_t
% &= \mathbf{H}_0 + f_1^{\mathrm{RF}}(\mathbf{H}_0)
% + f_2^{\mathrm{RF}}\!\left(f_1^{\mathrm{RF}}(\mathbf{H}_0)\right) + \cdots
% + f_L^{\mathrm{RF}}\!\left(f_{L-1}^{\mathrm{RF}}\!\left(\cdots f_1^{\mathrm{RF}}(\mathbf{H}_0)\cdots\right)\right).
% \end{aligned}
% \end{equation}
% In contrast to the residual accumulation in \cref{eq:residual-sum}, this fusion is performed only when constructing the encoder output. This allows features from shallower blocks to remain accessible to the decoder without requiring them to be retained in deeper blocks.

\paragraph{Attention initialization.}
Self-attention is disproportionately ill-conditioned compared with
other components of a transformer block~\citep{ji2025always},
and residual connections are commonly used to stabilize its training.
To improve its conditioning and convergence in the absence of
residual connections, one approach is mimetic
initialization~\citep{trockman2023mimetic,ji2026cutting},
which initializes the query and key projections to satisfy
\begin{equation}
    (\mathbf{W}_{\mathrm{Q}}^\ell)
    (\mathbf{W}_{\mathrm{K}}^\ell)^\top
    = \alpha\mathbf{Z}+\beta\mathbf{I},
\end{equation}
where $\mathbf{Z}_{ij}\sim\mathcal{N}(0,1/D)$,
$\mathbf{I}$ is the identity matrix, and $\alpha$ and $\beta$
are scalar constants.
This initialization encourages diagonal dominance in the query--key
product, which can improve the conditioning of self-attention
and support stable and fast convergence.

However, applying this approach to new models involves tuning
two hyperparameters, $\alpha$ and $\beta$.
We therefore propose a new strategy that mixes a shared orthogonal
component with the default query and key initializations.
For each encoder block $\ell$, we sample a random orthogonal matrix
$\mathbf{O}_\ell\in\mathbb{R}^{D\times D}$ and set
\begin{equation}
\begin{aligned}
\mathbf{W}_{\mathrm{Q}}^\ell
&\leftarrow
\sqrt{1-\rho}\,\mathbf{O}_\ell
+\sqrt{\rho}\,\mathbf{W}_{\mathrm{Q,default}}^\ell,\\
\mathbf{W}_{\mathrm{K}}^\ell
&\leftarrow
\sqrt{1-\rho}\,\mathbf{O}_\ell
+\sqrt{\rho}\,\mathbf{W}_{\mathrm{K,default}}^\ell,
\end{aligned}
\label{eq:qk-mix-init}
\end{equation}
The entries of $\mathbf{W}_{\mathrm{Q,default}}^\ell$
and $\mathbf{W}_{\mathrm{K,default}}^\ell$ are sampled
independently from the uniform distribution
$\mathcal{U}(-1/\sqrt{D},\,1/\sqrt{D})$,
where $D$ is the input feature dimension.
The parameter $\rho \in [0,1]$ controls the contribution
of these random components relative to the shared
orthogonal component.
The orthogonal component is
shared between the query and key projections, whereas the default
components are drawn independently. At $\rho = 0$, both projections
start from the same orthogonal matrix with condition number
$\kappa = 1$. At $\rho = 1$, they retain the default
initialization. The projections are tied only at initialization and
trained as separate parameters. The value and output projections
$\mathbf{W}^{\ell}_{\mathrm{V}}, \mathbf{W}^{\ell}_{\mathrm{O}}$
are initialized as orthogonal matrices and thus have condition number
$\kappa = 1$.

\paragraph{MLP initialization.}
% Similarly, we initialize every MLP weight matrix in the encoder as a  semi-orthogonal matrix,similarly in ~\citet{martens2021rapid,ji2026quantization}, so
% that its it preserves the well-conditioning of
% MLP at initialization, as the value and output projections do in
% Self-Attention.

Using an approach similar to \citet{martens2021rapid} and \citet{he2023deep}, we initialize weight matrices in the MLP blocks according to the scale-corrected uniform orthogonal (SUO) distribution, which ensures that these layers preserve the vector RMS of their inputs, whilst also having a condition number of 1. For the depths we trained at, we found that it wasn't necessary to transform the activation functions as recommended in \citet{martens2021rapid}.  % such that it preserves the well-conditioning of MLP at initialization, as the value and output projections do in Self-Attention.

%%%%%%%%%%%%%%%%%%%%%%%%%%%%%%%%%%%%%%%%%%%%%%%%%%%%%%%%%%%%%%%%%%%%%%%%%%

%% file: sec/5_experiment.tex
\section{Experiments}
We evaluate our methods in training diffusion transformers from scratch on the ImageNet-1k dataset at a resolution of $256 \times 256$ and patch size of 2 for 80 epochs, using a total training batch size of 1024. Following~\citet{zhou2026guiding}, we use the Muon optimizer with a constant learning rate of 0.0002 and a momentum of 0.95 throughout training. We set the gradient clipping threshold to 1, weight decay to 0, and the exponential moving average (EMA) decay rate to 0.9999. We use the off-the-shelf VAE-ft-EMA from Hugging Face\footnote{https://huggingface.co/stabilityai/sd-vae-ft-ema}, with a downsampling factor of 8. For sampling, we use the Euler solver with 250 steps. We train both DDT and DDT-RFE in both Large (L) and Extra Large (XL) configurations, with 458M and 675M parameters, respectively. For DDT-RFE, we set $\rho=0.1$ for the L configuration and
$\rho=0$ for the XL configuration. The original DDT training setup~\citep{wang2026ddt} incorporates REPA~\citep{yu2025representation} to align intermediate encoder features with representations from a pretrained visual encoder. We omit this auxiliary alignment objective in our experiments to examine the encoder representations learned through the generative objective. We evaluate both representation learning tasks using the intermediate features learned by the diffusion transformer models and the generation performance of our trained models.

\subsection{Representation Learning}
To assess how our architectural design affects the representations learned by diffusion models, we evaluate representations on four visual tasks: image classification, object discovery, semantic correspondence and semantic segmentation. For all evaluations, we extract features using conditioning on the null class.

\subsubsection{Image Classification}
\begin{figure}
    \centering
    \includegraphics[width=0.95\linewidth]{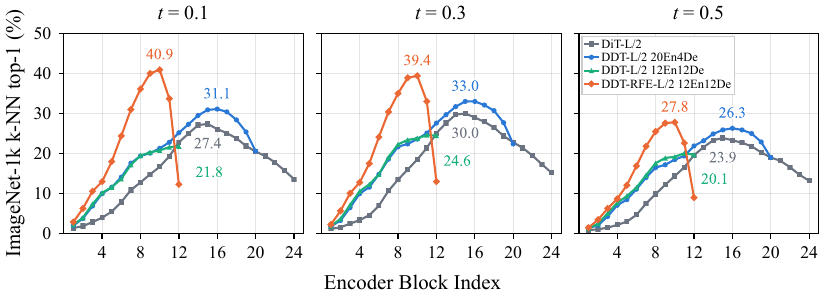}
    \caption{kNN top-1 accuracy on ImageNet-1k for the representation of every (encoder) block, at noise levels $t \in \{0.1, 0.3, 0.5\}$.}
    \label{fig:knn_in1k}
\end{figure}

\paragraph{Setting.}
We assess the representations of the frozen models by classification on ImageNet-1k. Given a noised latent $\mathbf{x}_t$, we extract token features from an encoder block and apply global average pooling to obtain an image-level representation. The same representation is used for both evaluations below. First, following the DINO protocol~\citep{caron2021emerging}, we evaluate $k$-nearest-neighbor (kNN) top-1 accuracy for every encoder block at noise levels $t \in \{0.1, 0.3, 0.5\}$, using the same timestep $t$ to construct the noisy input and to condition the network. Second, following DDAE~\citep{xiang2023denoising}, we evaluate linear separability by training a classification head consisting of BatchNorm and a linear layer for 30 epochs. For each model we select the encoder block and noise level with the best validation accuracy and report top-1 and top-5 accuracy, and we additionally probe the encoder output that conditions the decoder with the same procedure. We compare DiT-L, DDT-L in its best configuration (20en4de), and DDT-RFE-L (12en12de). Because DDT-RFE uses fewer encoder blocks than the best DDT configuration, we also train DDT-L with the same encoder depth as ours (12en12de) for a fair comparison.

\begin{table}
  \centering
  \caption{Linear probing accuracy (\%) on ImageNet-1k. We report the
    best over encoder blocks and noise levels $t$ ($t=0$ is
    the clean latent, $t=1$ pure noise).
    ``Encoder Output'' refers to the representation supplied to the decoder: the final
    encoder-block representation for the DDT, the aggregated encoder representation for
    the DDT-RFE.}
  \label{tab:probe_in1k}
  \resizebox{\linewidth}{!}{%
  \begin{tabular}{lcccrrcrr}
    \toprule
    & & \multicolumn{4}{c}{Best Encoder Block} & \multicolumn{3}{c}{Encoder Output} \\
    \cmidrule(lr){3-6}\cmidrule(lr){7-9}
    Model & Blocks (En/De) & Block & $t$ & top-1 & top-5 & $t$ & top-1 & top-5 \\
    \midrule
    DDT-L/2      & 20 / 4  & 15 & 0.15 & 59.6          & 82.1          & 0.20 & 51.8          & 75.7 \\
    DDT-RFE-L/2 (Ours)  & 12 / 12 &  8 & 0.05 & \textbf{60.9} & \textbf{83.5} & 0.15 & \textbf{59.7} & \textbf{82.4} \\
    \midrule
    DDT-XL/2     & 22 / 6  & 17 & 0.15 & 61.8          & 84.0          & 0.15 & 57.1          & 80.0 \\
    DDT-RFE-XL/2 (Ours) & 14 / 14 &  9 & 0.05 & \textbf{62.4} & \textbf{84.7} & 0.10 & \textbf{61.3} & \textbf{83.7} \\
    \bottomrule
  \end{tabular}
  }
\end{table}

\paragraph{Results and analysis.}
\Cref{fig:knn_in1k} reports kNN accuracy across blocks. As expected, DDT-L outperforms DiT-L at every noise level, reflecting the benefit of decoupling the encoder from the decoder. The DDT-L configuration with more encoder blocks also achieves higher accuracy. For example, at $t=0.1$, DDT-L (20en4de) reaches 31.1\%, compared with 21.8\% for DDT-L (12en12de). Despite using fewer encoder blocks, DDT-RFE-L improves over DDT-L by 9.8\% at $t=0.1$ and 6.4\% at $t=0.3$. The linear probing results in \cref{tab:probe_in1k} also show that DDT-RFE outperforms DDT in both top-1 and top-5 accuracy at the L and XL scales. At both scales, DDT-RFE achieves its best linear probing performance at a lower noise level ($t=0.05$ versus $t=0.15$ for DDT). Moreover, the drop in top-1 accuracy from the best encoder block to the encoder output is 7.8 and 4.7 percentage points for DDT-L and DDT-XL, respectively, compared with only 1.2 and 1.1 percentage points for DDT-RFE-L and DDT-RFE-XL, respectively. The encoder output of DDT-RFE thus remains nearly as semantically discriminative as its best intermediate representations for the classification task, whereas DDT exhibits a more pronounced gap between the two. DDT-RFE achieves these improvements with substantially fewer encoder blocks (12 versus 20 at the L scale and 14 versus 22 at the XL scale), corresponding to reductions of 40.0\% and 36.4\% in encoder depth, respectively.

\subsubsection{Object Discovery}
\paragraph{Setting.}
Detecting salient objects is a fundamental problem in computer vision with applications in real-world vision systems. To address this problem, recent methods, such as LOST~\citep{simeoni2021localizing} and TokenCuT~\citep{wang2023tokencut}, use self-supervised pretrained models to obtain high-quality, abstract feature representations without requiring manual labels. We evaluate unsupervised object discovery using LOST, which localizes the salient object in an image using token features from a frozen model. We apply LOST to representations extracted from each transformer block and report the best results. We evaluate on VOC2012 and COCO20k and report correct localization (CorLoc), defined as the fraction of images for which the predicted bounding box has an intersection-over-union (IoU) greater than 0.5 with a ground-truth bounding box.

\begin{table}[H]
  \centering
  \caption{Unsupervised object discovery with LOST
    on frozen encoders: CorLoc (\%) on VOC2012 and COCO20k. We report the best over encoder blocks and noise levels.}
  \label{tab:lost}
  \begin{tabular}{lcrr}
    \toprule
    Model & Blocks (En/De) & VOC2012 & COCO20k \\
    \midrule
    DDT-L/2      & 20 / 4  & 42.5          & 31.1 \\
    DDT-RFE-L/2 (Ours)  & 12 / 12 & \textbf{58.4} & \textbf{41.4} \\
    \midrule
    DDT-XL/2     & 22 / 6  & 41.3          & 29.8 \\
    DDT-RFE-XL/2 (Ours) & 14 / 14 & \textbf{57.8} & \textbf{39.0} \\
    \bottomrule
  \end{tabular}
\end{table}

\paragraph{Results and analysis.}
\Cref{tab:lost} reports CorLoc results. At both model scales, DDT-RFE improves over DDT by about 16 \% on VOC2012 and 10 \% on COCO20k. LOST relies on a clear separation between foreground and background features. When features retain too much low-level information from shallow blocks, patch similarities may reflect local appearance more than object structure. This can make it harder to select a foreground seed and identify the full object. With the residual-free encoder, tokens from the same object have more similar features, as also shown in \cref{fig:teaser}. These observations suggest that DDT-RFE learns features that better capture object semantics and distinguish objects from the background.

\subsubsection{Semantic correspondence}
\paragraph{Setting.}
Semantic correspondence aims to match points on corresponding parts of different instances of the same object category. These instances may differ in appearance, pose, viewpoint, and scale. This task tests whether local features capture consistent semantic information across images and support accurate localization of object parts. We evaluate zero-shot semantic correspondence on SPair-71k~\citep{min2019spair}, which is the most challenging semantic correspondence dataset,
containing diverse variations in viewpoint and scale with 12,234 image pairs on 18 categories for
testing, following the protocol of DIFT~\citep{tang2023emergent}. We report PCK@0.1$_{\text{bbox}}$, the percentage of keypoints whose predicted matches lie within $0.1\times\max(w,h)$ of their ground-truth locations, where $w$ and $h$ are the width and height of the target object's bounding box.

\begin{table}[t]
  \centering
  \caption{Zero-shot semantic correspondence on SPair-71k (PCK@0.1$_{\text{bbox}}$,
    \%), following the DIFT protocol at $256\times256$ input with the features
    of one encoder block. No training is involved; we report the best over
    encoder blocks and noise levels.}
  \label{tab:corr}
  \begin{tabular}{lcr}
    \toprule
    Model & Blocks (En/De) & SPair-71k \\
    \midrule
    DDT-L/2      & 20 / 4  & 33.3 \\
    DDT-RFE-L/2 (Ours)  & 12 / 12 & \textbf{37.2} \\
    \midrule
    DDT-XL/2     & 22 / 6  & 35.2 \\
    DDT-RFE-XL/2 (Ours) & 14 / 14 & \textbf{39.9} \\
    \bottomrule
  \end{tabular}
\end{table}
\paragraph{Results and analysis.}
\Cref{tab:corr} shows that DDT-RFE improves PCK over DDT by 3.9 and 4.7 percentage points at the L and XL scales, respectively. These gains are achieved by directly matching frozen features, without training a task-specific prediction head. They suggest that DDT-RFE learns features that more reliably identify corresponding object parts across different instances. This complements the object discovery results that the features support both locating whole objects and matching their parts across images. DDT-RFE achieves these improvements with fewer encoder blocks, using 12 versus 20 at the L scale and 14 versus 22 at the XL scale.

\subsubsection{Semantic Segmentation}
\paragraph{Setting.}

We evaluate linear probing on intermediate features for the semantic segmentation task. A classifier is trained on top of the representations, with performance measured by mean intersection-over-union (mIoU). Our linear probing protocol largely follows DDAE++~\citep{xiang2025ddae++} and sweep over all best blocks and noise levels. We evaluate both single representation and multiscale representations from different blocks. We report results on VOC2012~\citep{everingham2015pascal}, ADE20K~\citep{zhou2019semantic} and COCO-Stuff-10K~\citep{caesar2018coco}.

\paragraph{Results and analysis.}
\Cref{tab:seg} reports mIoU on the three datasets. On VOC2012, DDT-RFE outperforms DDT by about 6 mIoU at both L and XL scales, with single-block and with multi-scale features. On ADE20K and COCO-Stuff, the single-block gains are smaller. At the XL scale, DDT-RFE matches DDT on COCO-Stuff and is 0.3 mIoU below it on ADE20K. We attribute this to the greater scene complexity of these datasets, where multi-scale information is critical. The residual connections of DDT mix representations implicitly across depth, so a single block already carries multi-level information. However, the residual-free encoder carries more abstract but less multi-scale features. Multi-scale probing is consistent with this and it benefits DDT-RFE more than DDT on both scene datasets (for example, +2.9 versus +1.8 mIoU on ADE20K at the L scale), and with multi-scale features DDT-RFE leads on every dataset at both scales. In the residual-free encoder, abstraction arises within the hierarchy and multi-scale context is recovered by aggregation. DDT-RFE achieves these results with a shallower encoder than DDT, obtaining better semantic representations at lower cost.
\begin{table}
  \centering
  \caption{Dense linear probing for semantic segmentation (mIoU) on VOC2012, ADE20K and
    COCO-Stuff 10K.
    We report the
    best over encoder blocks and noise levels with single and multiscale representations.}
  \label{tab:seg}
  \begin{tabular}{lcrrrrrr}
    \toprule
    & &  \multicolumn{2}{c}{VOC2012} & \multicolumn{2}{c}{ADE20K} & \multicolumn{2}{c}{COCO-Stuff} \\
    \cmidrule(lr){3-4}\cmidrule(lr){5-6}\cmidrule(lr){7-8}
    Model & Blocks (En/De)  & single & multi & single & multi & single & multi \\
    \midrule
    DDT-L/2  & 20 / 4        & 50.5          & 54.2          & 25.8          & 27.6          & 25.5          & 26.7 \\
    DDT-RFE-L/2 (Ours)  & 12 / 12  & \textbf{57.0} & \textbf{60.5} & \textbf{26.8} & \textbf{29.7} & \textbf{26.0} & \textbf{28.2} \\
    \midrule
    DDT-XL/2 & 22 / 6        & 53.8          & 56.9          & \textbf{27.9} & 29.8          & \textbf{27.0} & 28.2 \\
    DDT-RFE-XL/2 (Ours) & 14 / 14  & \textbf{59.8} & \textbf{63.1} & 27.6          & \textbf{30.0} & \textbf{27.0}          & \textbf{29.0} \\
    \bottomrule
  \end{tabular}
\end{table}

\subsection{Image Generation}
\paragraph{Setting.}
We evaluate class-conditional image generation on ImageNet at $256\times256$ resolution to assess whether the representation improvements are accompanied by better generation performance. For each model trained for 80 epochs, we generate 50,000 samples using EMA weights and a 250-step Euler ODE sampler, without classifier-free guidance. We report Fr\'echet Inception Distance (FID), spatial FID (sFID), Inception Score (IS), precision, and recall.

\begin{table}[t]
  \centering
  \caption{Benchmarking class-conditional image generation on ImageNet $256{\times}256$ generation. We use 50k samples, 250-step Euler ODE
    sampler without classifier-free guidance. We report FID, IS, sFID, Precision and Recall.}

  \label{tab:gen_ep80}
  \begin{tabular}{lcrrrrr}
    \toprule
    Model & Blocks (En/De) & FID$\downarrow$ & IS$\uparrow$ & sFID$\downarrow$ & Prec.$\uparrow$ & Rec.$\uparrow$ \\
    \midrule
    DiT-L/2      & 24 (--)  & 9.34          & 111.6          & \textbf{4.63} & 0.699          & \textbf{0.634} \\
    DDT-L/2      & 20 / 4   & 8.98          & 112.9          & 5.37          & 0.708          & 0.628 \\
    DDT-L/2      & 12 / 12  & 8.82          & 108.8          & 4.80          & 0.712          & 0.626 \\
    DDT-RFE-L/2 (Ours)  & 12 / 12  & \textbf{7.68} & \textbf{118.4} & 5.37          & \textbf{0.731} & 0.615 \\
    \midrule
    DDT-XL/2     & 22 / 6   & 7.27          & \textbf{125.6} & 5.89 & 0.724          & \textbf{0.624} \\
    DDT-RFE-XL/2 (Ours) & 14 / 14  & \textbf{6.81} & \textbf{125.6}          & \textbf{5.50}          & \textbf{0.740} & 0.610 \\
    \bottomrule
  \end{tabular}
\end{table}
\paragraph{Results and analysis.}
\Cref{tab:gen_ep80} shows that DDT-RFE achieves the lowest FID among the compared models at both scales. DDT-RFE-L reaches an FID of 7.68, compared with 8.98 for DDT-L and 9.34 for DiT-L. The improvement remains when DDT-L uses the same encoder and decoder depths as DDT-RFE-L. At the XL scale, DDT-RFE outperforms DDT in terms of FID (6.81 vs. 7.27), sFID (5.50 vs. 5.89), and precision (0.740 vs. 0.724). The two models achieve the same IS, although DDT-RFE has slightly lower recall. Overall, the gains in representation quality are accompanied by improved FID at both scales under the same training budget. We demonstrate generated images by our model in ~\cref{fig:generation_samples}

%% file: sec/6_conclusion.tex
\section{Limitations and Future Work}
Due to limited computational resources, our experiments focus on DDT models trained on ImageNet at $256\times256$ resolution for 80 epochs. We leave experiments with longer training schedules, higher resolutions (e.g., $512\times512$), and other diffusion architectures to future work to assess the broader applicability of our approach.

\section{Conclusion}
In this work, we studied how residual connections in the condition encoder shape the representations learned by diffusion transformers, with a focus on how the encoder learns abstract features.
We introduced DDT-RFE, which combines a residual-free encoder with
feature aggregation at its output. This design allows successive
encoder blocks to transform features without identity shortcuts,
while keeping information from different encoder depths accessible
to the decoder. With the proposed initialization, DDT-RFE achieves
overall improvements over DDT across four visual understanding tasks
using fewer encoder blocks. It also achieves a lower FID for image
generation on ImageNet under the same training budget. Our ablations
further show the importance of both initialization and output feature
aggregation. Together, these results suggest that separating feature
transformation within the encoder from aggregation at its output can
benefit both representation learning and image generation. We hope that our findings encourage the community to reconsider how information flows between blocks and how this affects the learning of abstract representations in diffusion models.

%% file: sec/7_appendix.tex
\subsection*{AI use statement}
In this work, we used generative AI tools to polish the language of the manuscript and improve its clarity and readability, implement and edit Python code for the proposed method, and create or modify figures and tables. We take responsibility for the final content of this work.

\appendix
\section{Appendix}

\subsection{Generation Image}
\begin{figure}[H]
    \centering
    \includegraphics[width=\linewidth,height=0.95\textheight,keepaspectratio]{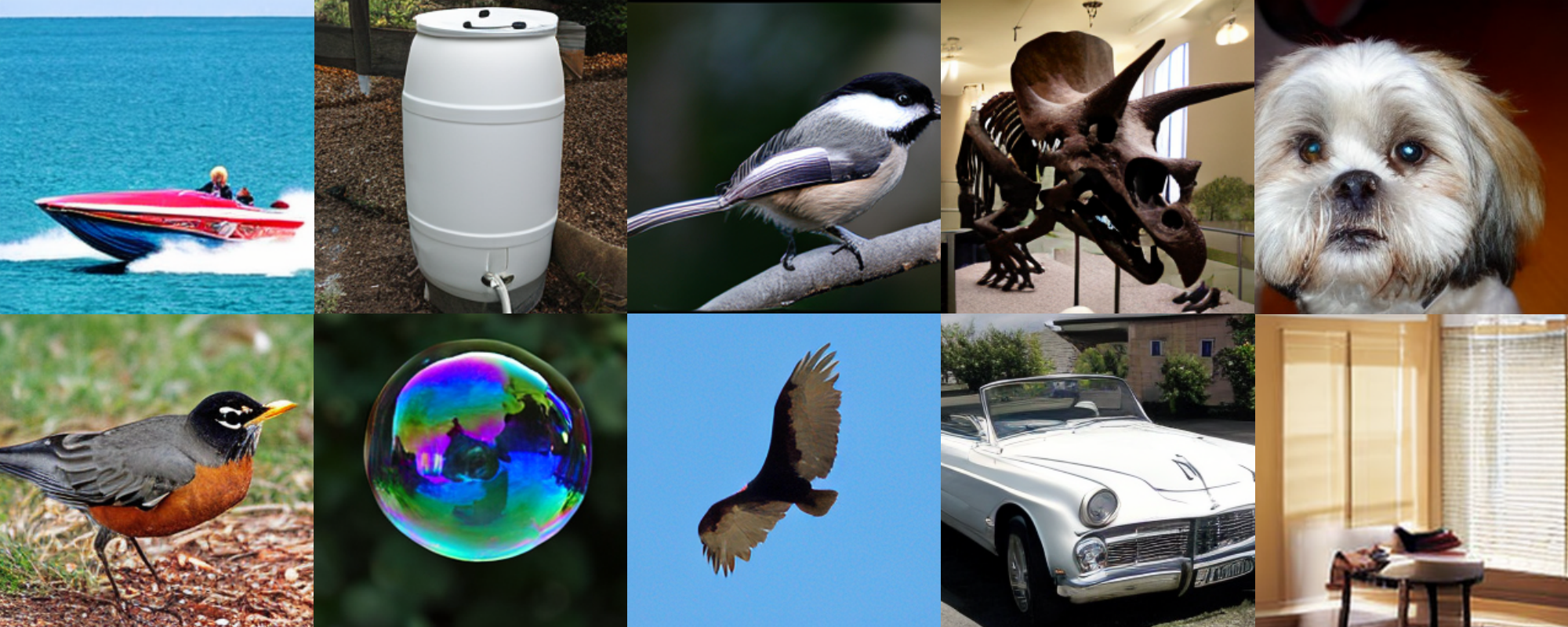}
    \caption{We use a 250-step Euler ODE sampler with cfg=4.}
    \label{fig:generation_samples}
\end{figure}

\Cref{fig:generation_samples} shows images generated by DDT-RFE-L/2 on ImageNet at $256\times256$ resolution.

\subsection{Self-Attention and MLP Operations}
\label{app:sa-mlp}
We detail the operations in \cref{eq:sublayers}, following the Self-Attention
formulation of \citet{ji2026cutting}.
Let $\mathbf{H}\in\mathbb{R}^{N\times D}$ denote a generic input token matrix.
In each encoder block, the Self-Attention and MLP operation take input
$\mathbf{H}_{\ell-1}$ and $\tilde{\mathbf{H}}_\ell$, respectively,
with normalization and adaptive modulation applied as defined
in \cref{eq:sublayers}.
For clarity, we omit normalization layers, rotary positional embeddings
(RoPE), and bias terms, and suppress the block index $\ell$.

\paragraph{Self-attention.}
Let $n_h$ be the number of heads and $d_h=D/n_h$ the dimension of each
head. The query, key, and value matrices for head $h$ are
\begin{equation}
\mathbf{Q}_h=\mathbf{H}\mathbf{W}_{\mathrm{Q},h}^{\top},
\qquad
\mathbf{K}_h=\mathbf{H}\mathbf{W}_{\mathrm{K},h}^{\top},
\qquad
\mathbf{V}_h=\mathbf{H}\mathbf{W}_{\mathrm{V},h}^{\top},
\label{eq:app-qkv}
\end{equation}
where $\mathbf{W}_{\mathrm{Q},h},\mathbf{W}_{\mathrm{K},h},
\mathbf{W}_{\mathrm{V},h}\in\mathbb{R}^{d_h\times D}$.
The attention matrix is
\begin{equation}
\mathbf{A}_h=\eta\!\left(\mathbf{Q}_h\mathbf{K}_h^{\top}\right),
\qquad
\eta(\mathbf{S})=\mathrm{softmax}\!\left(
\frac{\mathbf{S}}{\sqrt{d_h}}\right),
\label{eq:app-attention-matrix}
\end{equation}

The head outputs are concatenated and projected as Self-Attention output:
\begin{equation}
\mathrm{SA}(\mathbf{H})
=\mathrm{Concat}\!\left(
\mathbf{A}_1\mathbf{V}_1,\ldots,
\mathbf{A}_{n_h}\mathbf{V}_{n_h}
\right)\mathbf{W}_{\mathrm{O}}^{\top},
\label{eq:app-sa}
\end{equation}
where $\mathbf{W}_{\mathrm{O}}\in\mathbb{R}^{D\times D}$.

\paragraph{MLP.}
We use a two-layer MLP with a GELU activation, applied independently
to each token:
\begin{equation}
\mathrm{MLP}(\mathbf{H})
=\mathrm{GELU}\!\left(\mathbf{H}\mathbf{W}_1^{\top}\right)
\mathbf{W}_2^{\top},
\label{eq:app-mlp}
\end{equation}
where $\mathbf{W}_1\in\mathbb{R}^{D_{\mathrm{ff}}\times D}$,
$\mathbf{W}_2\in\mathbb{R}^{D\times D_{\mathrm{ff}}}$,
and $D_{\mathrm{ff}}$ is the hidden dimension that commonly is set to be $4D$.
The GELU activation is applied element-wise.

\subsection{Ablation on DDT}
To determine the relative importance of the difference model design choices, with a particular focus on the removal of residual connections, we train the DDT baseline with our proposed initialization and linear feature fusion for 80 epochs. \Cref{tab:abl_fusion_gen_cls} shows that applying these techniques together degrades generation performance while yielding comparable classification results. \Cref{tab:abl_fusion_dense} further shows that this variant performs worse on semantic segmentation and object discovery.
\begin{table}[H]
  \centering
  \caption{Performance on generation and classification tasks.}
  \label{tab:abl_fusion_gen_cls}
  \begin{tabular}{lcrrrr}
    \toprule
    & & \multicolumn{2}{c}{Generation} & \multicolumn{2}{c}{Classification} \\
    \cmidrule(lr){3-4}\cmidrule(lr){5-6}
    & & \multicolumn{2}{c}{ImageNet $256{\times}256$} & \multicolumn{2}{c}{Best encoder block} \\
    \cmidrule(lr){3-4}\cmidrule(lr){5-6}
    Model & Blocks (En/De) & FID$\downarrow$ & IS$\uparrow$ & top-1 & top-5 \\
    \midrule
    DDT-L/2             & 20 / 4  & 8.98 & \textbf{112.9} & 59.6          & 82.1 \\
    DDT-L/2             & 12 / 12 & \textbf{8.82}          & 108.8          & 51.0          & 75.2 \\
    DDT-L/2+linear+init & 20 / 4  & 10.90         & 99.3           & \textbf{60.4} & \textbf{82.8} \\
    \bottomrule
  \end{tabular}
\end{table}

\begin{table}[H]
  \centering
  \caption{Performance on Semantic segmentation and Object discovery.}
  \label{tab:abl_fusion_dense}
  \resizebox{\linewidth}{!}{%
  \begin{tabular}{lcrrrrrrrr}
    \toprule
    & & \multicolumn{6}{c}{Semantic segmentation (mIoU)} & \multicolumn{2}{c}{Object discovery} \\
    \cmidrule(lr){3-8}\cmidrule(lr){9-10}
    & & \multicolumn{2}{c}{VOC2012} & \multicolumn{2}{c}{ADE20K} & \multicolumn{2}{c}{COCO-Stuff} & VOC2012 & COCO20k \\
    \cmidrule(lr){3-4}\cmidrule(lr){5-6}\cmidrule(lr){7-8}\cmidrule(lr){9-9}\cmidrule(lr){10-10}
    Model & Blocks (En/De) & single & multi & single & multi & single & multi & & \\
    \midrule
    DDT-L/2             & 20 / 4  & \textbf{50.5} & \textbf{54.2} & \textbf{25.8} & \textbf{27.6} & \textbf{25.5} & \textbf{26.7} & \textbf{42.5} & \textbf{31.1} \\
    DDT-L/2             & 12 / 12 & 39.0          & 42.0          & 20.0          & 21.3          & 20.1          & 21.2          & 42.1          & 29.8 \\
    DDT-L/2+linear+init & 20 / 4  & 43.1          & 47.9          & 23.1          & 25.8          & 22.9          & 24.8          & 42.1          & 29.1 \\
    \bottomrule
  \end{tabular}
  }
\end{table}

\subsection{Ablation on DDT-RFE}

\begin{figure}
    \centering
    \includegraphics[width=\linewidth,height=0.95\textheight,keepaspectratio]{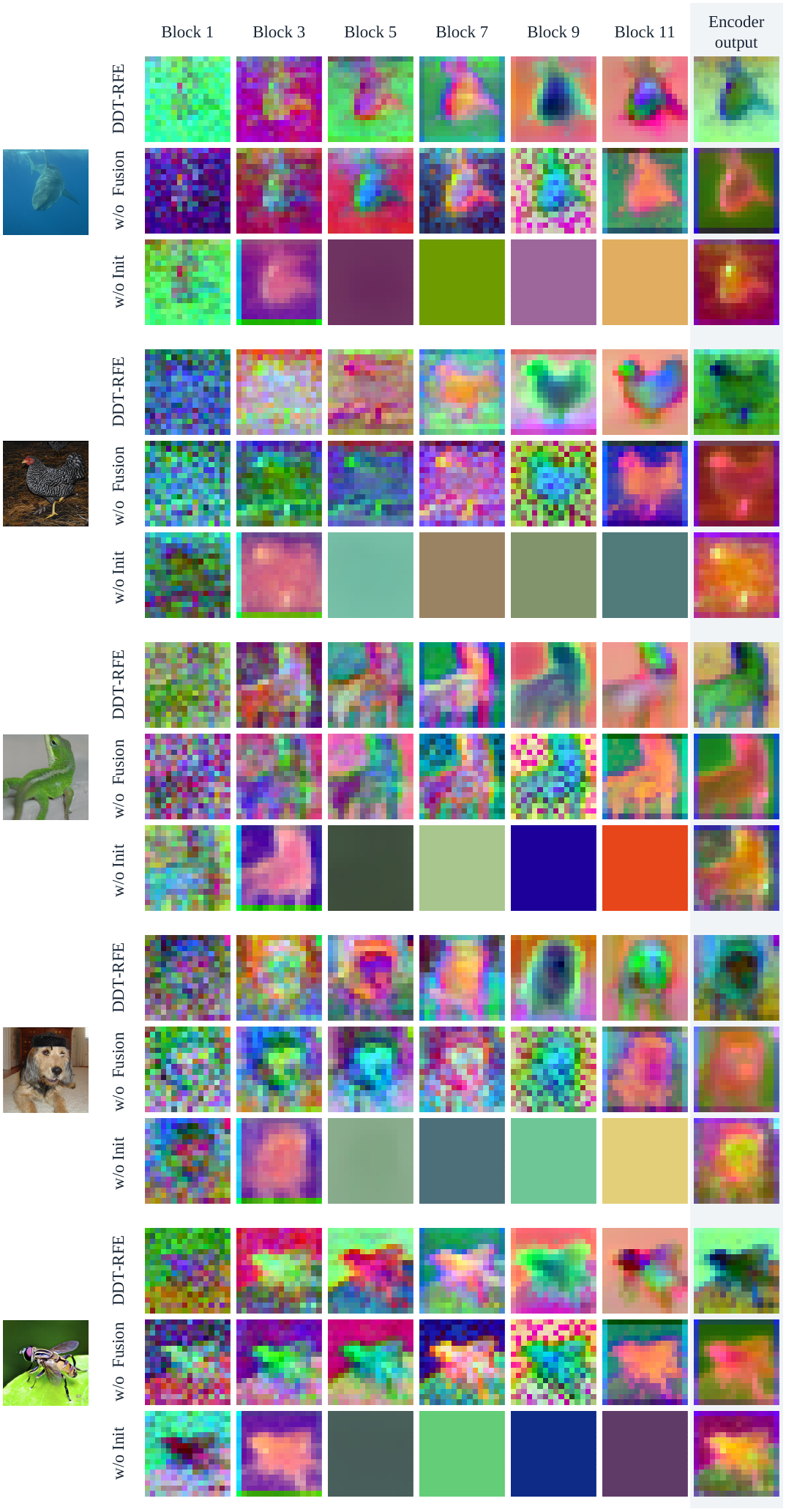}
    \caption{PCA visualization of DDT-RFE and its variants at noise level $t=0.5$.}
    \label{fig:ddt-rfe_pca}
\end{figure}

We ablate the proposed initialization scheme and feature fusion in DDT-RFE-L/2, with quantitative results in \cref{tab:abl_skipless} and feature visualizations in \cref{fig:ddt-rfe_pca}.
Removing the proposed initialization increases FID from 7.68 to 9.86 and substantially reduces performance on the visual understanding tasks. In \cref{fig:ddt-rfe_pca}, we can see that most representations have collapsed due to network conditioning issues~\citep{ji2025always}. Removing feature fusion only leads to much larger performance drops in both image generation and visual understanding. However, we empirically find that additionally removing the output gates helps recover performance. Nevertheless, this variant still underperforms the full recipe. Specifically, it increases FID by 0.66 and reduces classification accuracy, segmentation mIoU, and correspondence PCK by 0.2 percentage points, 1.1 points, and 4.4 percentage points, respectively. Qualitatively, the model learns some abstraction in the middle blocks, but it still attempts to retain detailed information because no alternative pathways are available.

\begin{table}[H]
  \centering
  \caption{Ablation study on DDT-RFE-L/2.}
  \label{tab:abl_skipless}
  \resizebox{\linewidth}{!}{%
  \begin{tabular}{lrrrrr}
    \toprule
    & \multicolumn{2}{c}{Generation} & Classification & Segmentation & Correspondence \\
    \cmidrule(lr){2-3}\cmidrule(lr){4-4}\cmidrule(lr){5-5}\cmidrule(lr){6-6}
    & \multicolumn{2}{c}{ImageNet} & ImageNet-1k & VOC2012 & SPair-71k \\
    \cmidrule(lr){2-3}\cmidrule(lr){4-4}\cmidrule(lr){5-5}\cmidrule(lr){6-6}
    & FID$\downarrow$ & IS$\uparrow$ & top-1 & mIoU & PCK@0.1 \\
    \midrule
    DDT-RFE-L/2 & \textbf{7.68} & \textbf{118.4} & \textbf{60.9} & \textbf{57.02} & \textbf{37.24} \\
    \quad \textbf{w/o} init & 9.86 & 105.7 & 38.6 & 30.62 & 13.89 \\
    \quad \textbf{w/o} fusion & 16.81 & 74.2 & 6.7 & 3.43 & 4.20 \\
    \quad \textbf{w/o} fusion, w/o gate & 8.34 & 114.8 & 60.7 & 55.91 & 32.87 \\
    \bottomrule
  \end{tabular}%
  }
\end{table}

\subsection{More Visualization}
\begin{figure}[H]
    \centering
\includegraphics[width=0.85\linewidth,height=0.85\textheight,keepaspectratio]{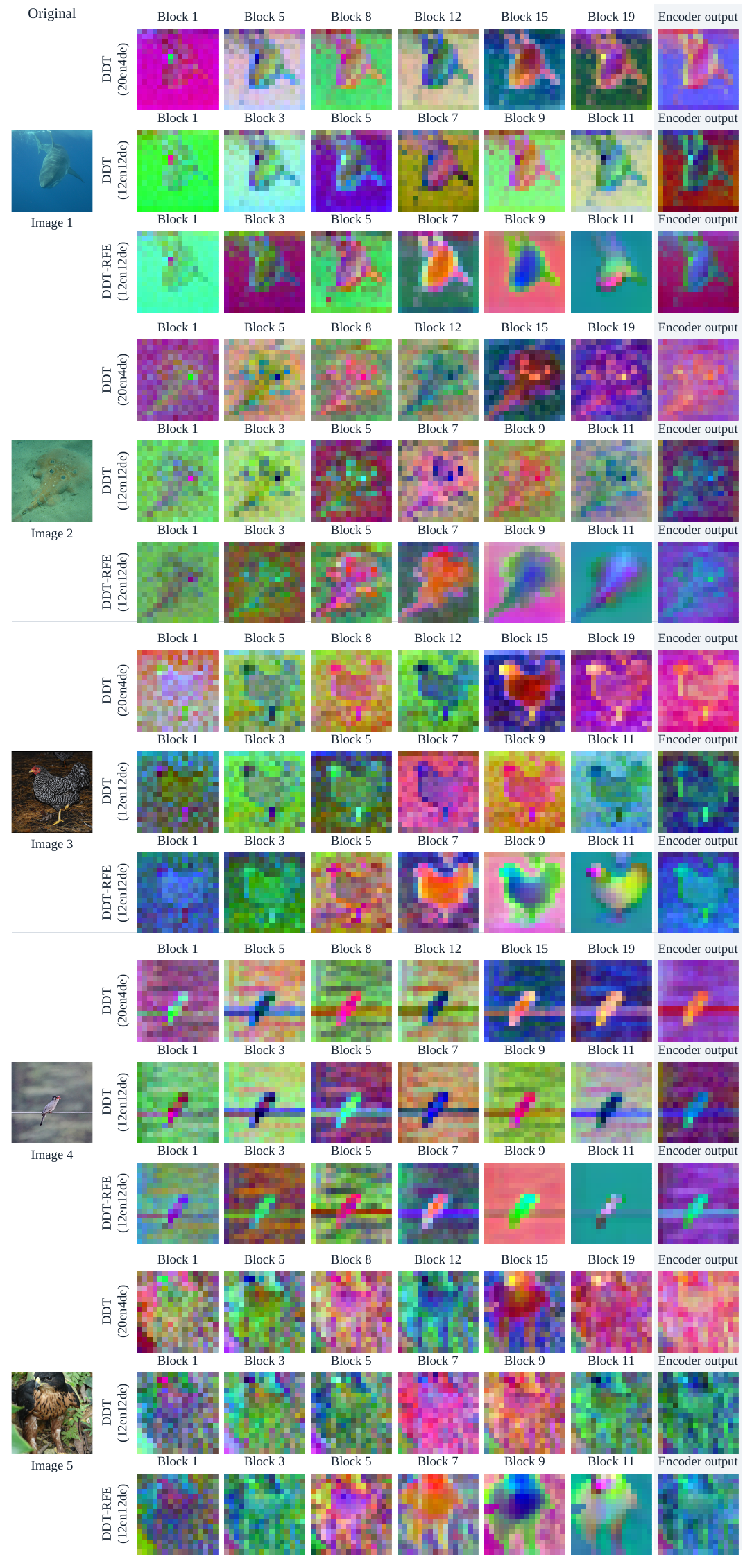}
    \caption{PCA visualization of DDT and DDT-RFE at noise level $t=0.1$}
    \label{fig:ddt_pca_t0.1}
\end{figure}

\begin{figure}
    \centering
    \includegraphics[width=\linewidth,height=0.95\textheight,keepaspectratio]{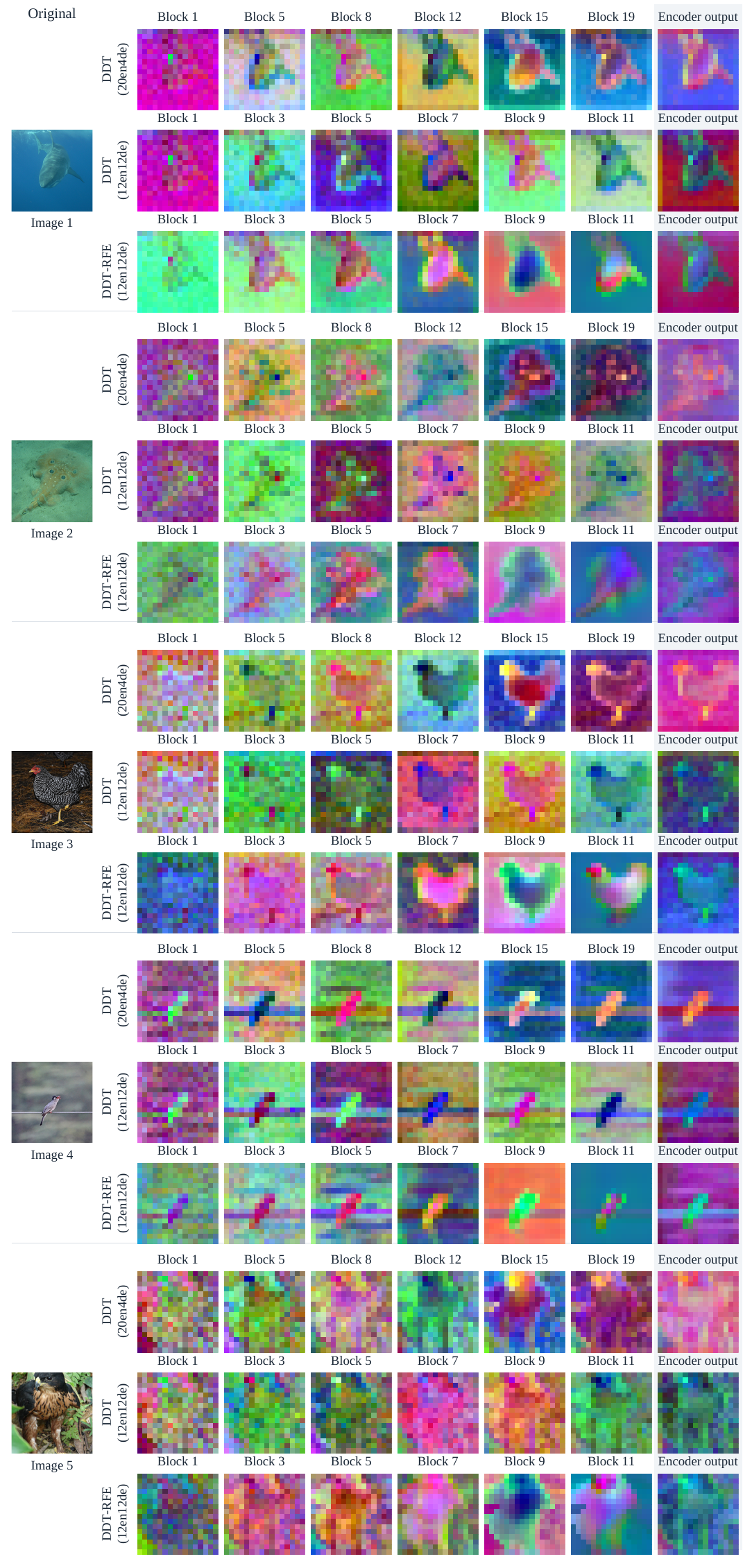}
    \caption{PCA visualization of DDT-RFE at noise level $t=0.3$}
    \label{fig:ddt_pca_t0.3}
\end{figure}

\begin{figure}
    \centering
    \includegraphics[width=\linewidth,height=0.95\textheight,keepaspectratio]{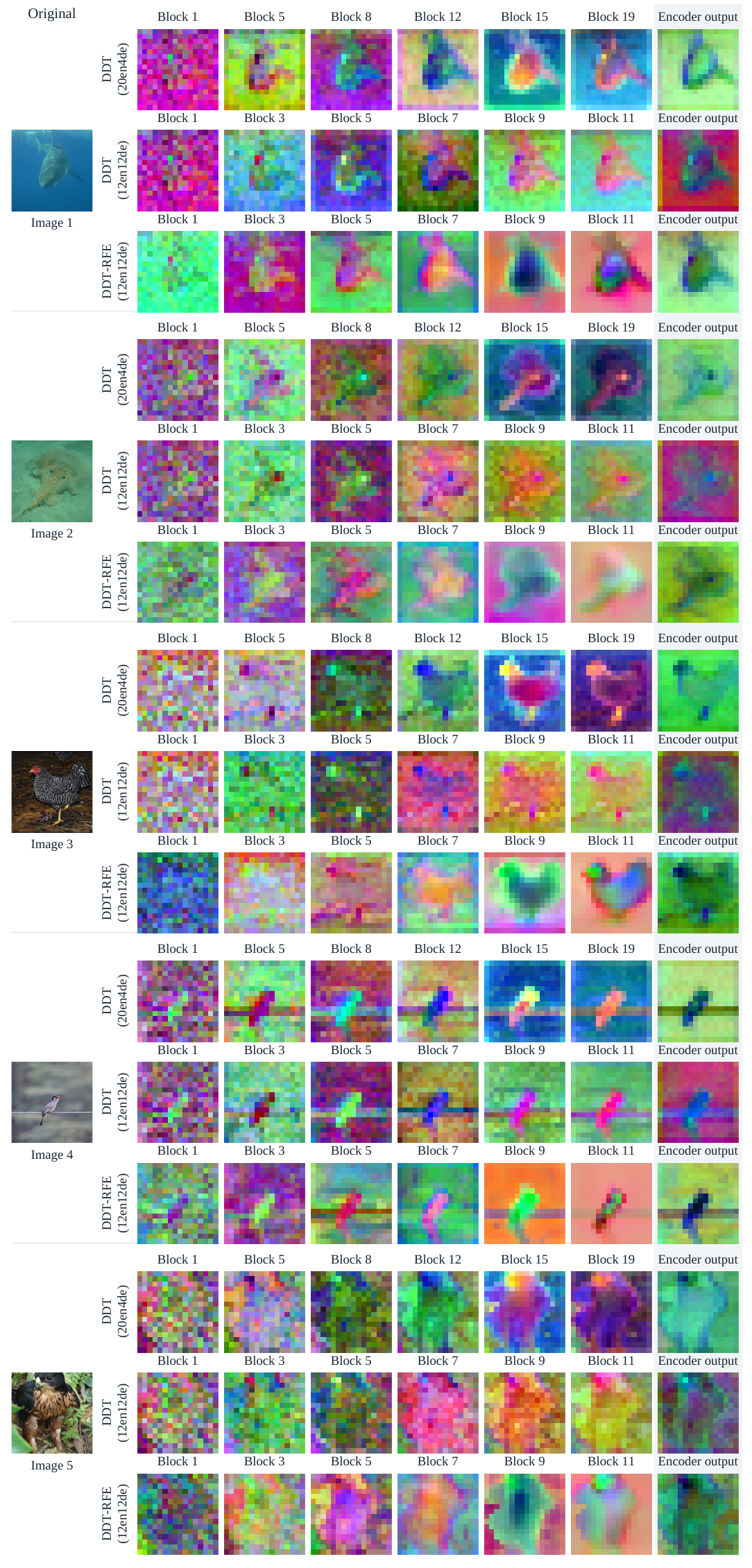}
    \caption{PCA visualization of DDT-RFE at noise level $t=0.5$}
    \label{fig:ddt_pca_t0.5}
\end{figure}

In this subsection, we demonstrate additional visualizations of encoder representations for three different models: DDT-L/2 (20en4de), DDT-L/2 (12en12de), and DDT-RFE-L/2 (12en12de) in ~\cref{fig:ddt_pca_t0.1,fig:ddt_pca_t0.3,fig:ddt_pca_t0.5}

